# A New Model of Plan Recognition

Robert P. Goldman, Christopher W. Geib and Christopher A. Miller
Honeywell Technology Center
Minneapolis, MN 55418 USA
{goldman,geib,cmiller}@htc.honeywell.com

## Abstract

We present a new abductive, probabilistic theory of plan recognition. This model differs from previous theories in being centered around a model of plan execution: most previous methods have been based on plans as formal objects or on rules describing the recognition process. We show that our new model accounts for phenomena omitted from most previous plan recognition theories: notably the cumulative effect of a sequence of observations of partially-ordered, interleaved plans and the effect of context on plan adoption. The model also supports inferences about the evolution of plan execution in situations where another agent intervenes in plan execution. This facility provides support for using plan recognition to build systems that will intelligently assist a user.

## 1 Introduction

In this paper, we present a new theory of plan recognition. Our theory, which is abductive and probabilistic, differs from previous theories in treating plan execution as primary. This new perspective clarifies a number of difficult issues in plan recognition, permitting solutions to a wider class of plan recognition problems and allows plan recognition work to proceed on a firmer theoretical footing.

In 1986, Kautz and Allen (K&A) published an article, "Generalized Plan Recognition," (Kautz & Allen 1986) that has framed the discussion of plan recognition ever since. K&A defined the problem of plan recognition as the problem of identifying a minimal set of *top-level actions* sufficient to explain the set of observed actions. Plans were represented in a plan graph, with top-level actions as root nodes and other actions as nodes depending from the top-level actions. To a first approximation, the problem of plan recognition was then a problem of graph covering. K&A formalized this view of plan recognition in terms of McCarthy's circumscription.

Unlike K&A's, our abductive, probabilistic theory is centered around plan *execution*, instead of around plan graphs as formal objects. Our new framework clarifies a number of issues that were obscured by previous approaches. In particular, our approach handles partially-ordered plans; multiple, interleaved plans; the effect of context on plan choice and is able to correctly treat a system that is both recognizing actions of other agents, and acting on its own right.

Like K&A, we offer only a formal model of the plan recognition problem. We do not pretend to provide a solution to either the algorithmic or implementation problems, only an understanding at the "knowledge level." We have encoded our plan recognition model in a language that has a computational interpretation, and we have a proof-of-concept interpreter that has been used to test all examples given in this paper. However, we do not claim that this interpreter provides an efficient solution to the plan recognition problem. There are many ways our model could be implemented; we mention several possibilities in this paper. For that matter, we do not believe that there is a single best algorithm, appropriate for all plan recognition domains.

Our work on plan recognition is motivated by an interest in procedure-based crisis management, specifically in industrial control. We are working on systems that will assist human agents executing crisis management procedures. As a result, our model of plan execution is hierarchical and procedural, as distinguished from models in which there are only primitive actions, or models that assume agents are deliberative (e.g., game-theoretic models). This application requires us to account for phenomena omitted from most previous plan recognition theories: notably the cumulative effect of a sequence of observations of partially-ordered, interleaved plans; the effect of context on plan adoption and the effect of interventions in the process of plan execution.

## 2   Plans

In this paper, we use simple hierarchical (task decomposition) plans, as most plan recognition work does. We assume that agents have a *plan library* that provides recipes for achieving goals. We give a sample plan library for a hypothetical space station example in Figure 1.

When an agent wishes to achieve a goal like increasing available power (**increase-power**), the plan library provides a set of alternate *methods* that the agent can use. **increase-power** can be achieved either by by generating more power (**gen-power**) or by reducing power consumption (**lower-power-use**). In order to **lower-power-use**, the agent must do three *steps*: open access panel 2 (**open-p2**), shutoff experiment sub-system X1 (**shutoff-X1**), and shutoff experiment sub-system X2 (**shutoff-X2**). Note that the plan library may be viewed as an AND/OR tree, with goals as OR nodes and methods as AND nodes.

Our space station plan library contains plans for three goals: increasing available power (**increase-power**), increasing the oxygen content of the air (**raise-$O_2$-level**) and raising the temperature (**raise-temp**). The goals **increase-power** and **raise-$O_2$-level** each have two alternate methods: generate more power (**gen-power**) and reduce power consumption (**lower-power-use**); and generate more $O_2$ (**gen-$O_2$**) and reduce $O_2$ consumption (**lower-$O_2$-use**), respectively. Since there is only one way of raising the temperature, there is no need for an actual method node for it.

Each of the method nodes has a set of children connected by "and" arcs representing the actions that are the steps of the method. In our simple plan library, the methods are all composed of primitive actions: open access panel 1 (**open-p1**), turn on generator B (**start-gen-B**), **open-p2**, **shutoff-X1**, **shutoff-X2**, start $O_2$ generator (**start-$O_2$-gen**), open access panel 3 (**open-p3**), seal-off science module (**seal-sci**), check temperature (**check-temp**), and raise thermostat setpoint (**raise-temp-set**). In general, however, methods may introduce sub-goals. For example, in an alternate library for this domain, **gen-$O_2$** might have **increase-power** as a sub-goal in place of the two actions **open-p1** and **start-gen-B**.

In many cases, the steps of a method must be done in a particular order. Ordering constraints are represented by directed arcs. For example, **open-p2** must precede **shutoff-X1** and **shutoff-X2** for **lower-power-use**.

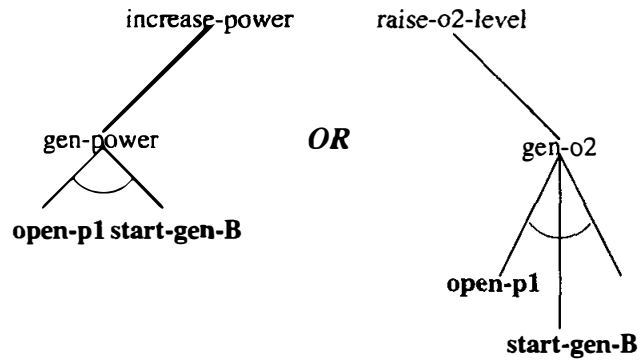

Figure 2: A minimal explanation involving only one plan.

Notice that the agent still has some freedom to choose the order of his/her actions.

Actions may be performed for more than one reason. For example, **open-p1** and **start-gen-B** are used in both **gen-power** and **gen-$O_2$**.

Finally, notice that there are two different conditions/events that are tied to the goals. These dashed lines represent the fact that these events affect the likelihood that the agent will adopt the respective goal. For example if an EVA is to be prepared (**EVA-prep**), then the agent is more likely to adopt **increase-power** as a goal. Likewise, if the agent notices a drop in $O_2$ level ($O_2$-drop), then he is likely to choose **raise-$O_2$-level** as a goal.

## 3   Plan Recognition

Plan recognition is the process of inferring the goals of an agent from observations of an agent's actions. Cohen, Perrault and Allen (1981) distinguish between two kinds of plan recognition, *keyhole* and *intended* plan recognition. In keyhole recognition, the recognizer is simply watching normal actions by the agent. In intended recognition, the agent is cooperative; its actions are done with the intent that they be understood. Keyhole recognition is the problem that faces us in our applications, and this fact influences the structure of our model.

To the best of our knowledge, Charniak was the first to argue that plan recognition was best understood as a specific form of the general problem of *abduction*, or reasoning to the best explanation (Charniak & McDermott 1985).

K&A's model of plan recognition (Kautz & Allen 1986) treated the problem as one of computing minimal explanations, in the form of vertex covers based on the plan graph. For example, in their approach, if one

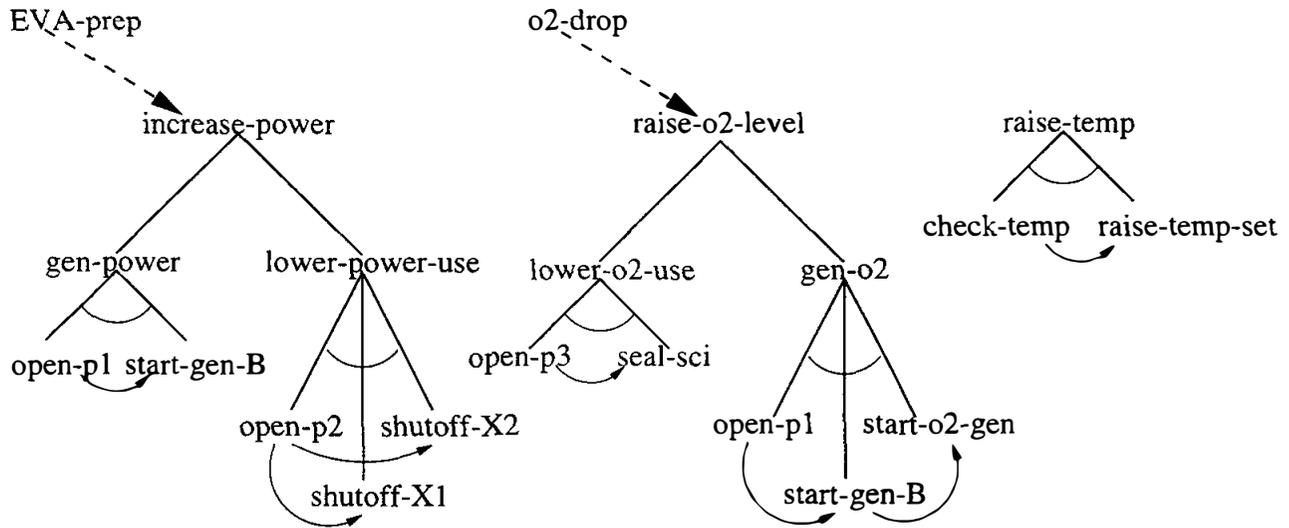

Figure 1: A hierarchical plan library in diagram form.

observed **open-p1** and **start-gen-B**, the minimal explanations would be

(**increase-power** ∧ **gen-power**) ∨
(**raise-$O_2$-level** ∧ **gen-$O_2$**)

See Figure 2. If, in addition, one observed **check-temp**, then the system would have to postulate *two* top-level plans in order to explain (cover) all the observations, as shown in Figure 3.

One problem with this work is that it does not take into account differences in the *a priori* likelihood of different plans. Charniak and Goldman (C&G) (1993) argued that, since plan recognition involves abduction, it could best be done as Bayesian inference. Bayesian inference supports the preference for minimal explanations, in the case of hypotheses that are equally likely, but also correctly handles explanations of the same complexity but different likelihoods.

Two plan recognition situations that are not handled by either K&A or C&G are the problems of influences from the state of the world and evidence from failure to observe. Clearly, the state of the world will influence an agent's decision to pursue plans. We see in our Figure 1 that the space station operator's decision to pursue **increase-power** will be affected substantially by whether or not an EVA is to be prepared (**EVA-prep**). K&A could not take this into account, because they did not treat the relative likelihood of plans. Even for C&G, however, it is not simple to take this into account, because they defined their probability distributions over the plan tree. On the other hand, when we view plan recognition from the point of view of agent's pursuing plans, it becomes clear how to handle this.

The problem of evidence from failure to observe is a more complex one. Consider what would happen if one observed **open-p1** and **start-gen-B**. Assuming that they were equally likely a priori, one would conclude that either **increase-power** (via **gen-power**) or **raise-$O_2$-level** (via **gen-$O_2$**) were equally good explanations (see Figure 2). However, as time went by and one saw other actions, without seeing **start-$O_2$-gen**, the final action of **gen-$O_2$**, one would become more and more certain that **increase-power/gen-power** was the right explanation. Systems like those of C&G and K&A, are not capable of reasoning like this, because they do not consider plan recognition as a problem that evolves over time. They cannot represent the fact that an action has not been observed *yet*. They can only be silent about whether an action has occurred — which just means that the system has failed to notice the action, not that the action hasn't occurred — or assert that an action has not *and will not* occur. We will show that our approach handles evidence from failure to observe actions.

Vilain (1990) presented a theory of plan recognition as parsing, based on K&A's theory. Vilain does not actually propose parsing as a solution to the plan recognition problem. Instead, he uses the reduction of limited cases of plan recognition to parsing in order to investigate the complexity of K&A's theory. The major problem with parsing as a model of plan recognition is that it does not treat partially-ordered plans or interleaved plans well. Indeed, even within a single method, partial ordering (as in **lower-power-use**, where **shutoff-X1** and **shutoff-X2** can be done in any order, as long as **open-p2** is done first), would cause an explosion in grammar size.

More recently, Wellman and Pynadath (W&P) (1997)

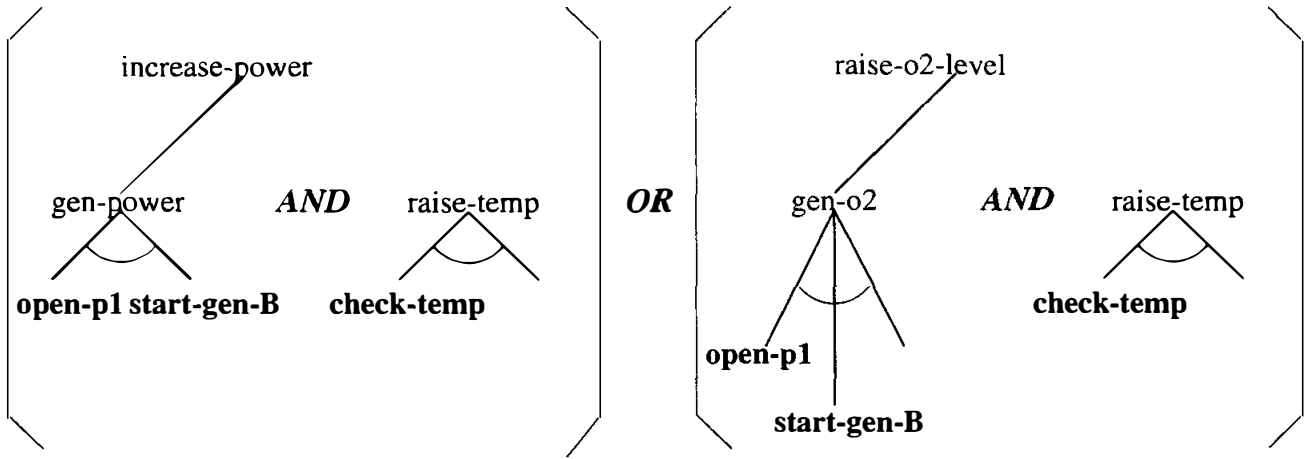

Figure 3: A minimal explanation for a situation that requires two plans.

have proposed a plan recognition method that is both probabilistic and based on parsing. Unfortunately, this approach suffers from the same limitations on plan interleaving as Vilain's. W&P propose that probabilistic context-*sensitive* grammars (PCSGs) might overcome this problem, but it is difficult to define a probability distribution for a PCSG (Pynadath & Wellman 1997). Difficulty in defining a probability distribution is also a problem for C&G's approach. Overcoming this difficulty is one of the emphases of our own work.

None of the plan recognition systems that we know of properly handles actions taken by the recognizing agent. How should the recognizing agent, A, reason about a situation in which it observes another agent, B, taking some actions, A infers a plan that it thinks B is pursuing, takes some actions on B's behalf, and then sees further actions by B? Judea Pearl has provided the general solution for reasoning about such situations, with his theory of interventions (Pearl 1994). We incorporate interventions into our theory of plan recognition, to permit recognizing systems to work with the agents whose plans they watch.

Shortcomings in past approaches motivated the work described in this paper. Previous plan recognition systems have had trouble handling:

- partially-ordered plans and plan interleaving;
- evidence from failure to observe actions;
- contextual influence on plan choice;
- domains in which the recognizer and the recognized both act.

In the following section, we will provide a novel model of plan recognition that provides solutions to these problems.

## 4 Plan execution model

Our model for plan execution is a simple one. The executing agent, at the start of the episode, chooses a set of plans to execute. The set of plans chosen determines the set of primitive actions that are *pending*. As the episode proceeds, the agent will repeatedly execute one of the pending actions, and generate a new set of pending actions from which further actions will be chosen. A new set of pending actions will be generated from the previous set by removing the action just executed and adding newly enabled actions. Actions become enabled when their predecessors are completed. This process is illustrated in Figure 5.

The model of plan execution provides a conceptual model for the *generation* of execution traces. In order to use this model to perform plan *recognition*, we use this model and reason abductively.

As a way of motivating some of the issues this work is designed to confront this section will present some brief examples of the kinds of problems our system is designed to handle.

### 4.1 Partial orderings

Most encodings of plan recognition raise the problem of partial orders. Partial orders arise through tasks that have sub-tasks that are only partially ordered. Method **lower-power-use** provides an example of partially-ordered sub-tasks: once the access panel has been opened, the two subsystems can be shut down in either order. Partial ordering also arises when an agent pursues multiple, independent tasks simultaneously. Of other plan recognition theories we have examined, only K&A's can properly handle partially-ordered tasks.

## 4.2 Overloaded actions

Many plan recognition systems do not permit explanations for an action in which that action is done as part of more than one plan. For example, **open-p1** and **start-gen-B** might be done as part of *both* the plans for **increase-power** *and* **raise-$O_2$-level**.

## 4.3 Performing actions for their own sake

Many previous plan recognition systems have required that their users make hard distinctions between "top-level" actions — actions that can be done for their own sake — and other actions that are done only in service of higher level goals. Our applications show that this distinction is artificial and unrealistic. For example, operators of control systems, when not otherwise occupied, often glance at the current values of process variables. We would expect, e.g., **check-temp** to be performed occasionally "for its own sake," rather than as part of **raise-temp**. Our model does not require a hard distinction between top-level and other actions.

## 4.4 Negative Evidence

Previous plan recognition systems have not been able properly to take into account negative evidence, the confirming or disconfirming effect of *failing* to see some action. Previous systems didn't take negative evidence into account because they didn't treat observations as sequences.

## 4.5 Context

Our model allows us to take into account the state of the world when considering what goals the agent might be pursuing. As we stated earlier, if the agent has seen a drop in $O_2$ level, then **raise-$O_2$-level** is a better explanation of **open-p1** than **increase-power** is.

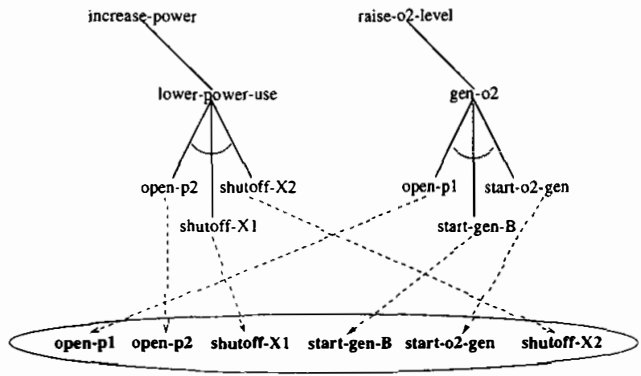

Figure 4: Explaining interleaved plans.

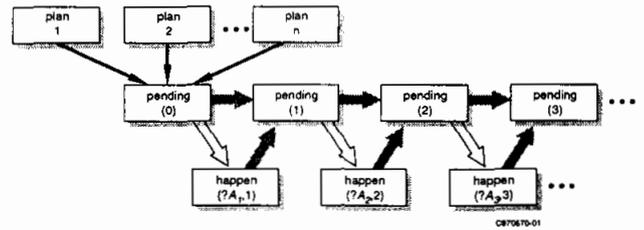

Figure 5: A dynamic belief net illustrating the simple model of plan execution.

After a more formal exposition of our model we will return to discuss these examples and the inference that is performed in our model.

## 5 Formalizing the model

In this section of the paper we present a formal representation of the plan execution model we have outlined above. As our notation we use Poole's logic of Probabilistic Horn Abduction (PHA) (Poole 1993b). PHA provides logical rules entailing propositions and distinguished assumable propositions called *hypotheses*. These rules are in Prolog-like Horn form. An explanation for a proposition is a set of hypotheses that, taken together with the rules, entail that proposition. Hypotheses have associated prior probabilities, so the logic supports a notion of best (most likely) explanation. Poole shows that PHA can be used to describe arbitrary belief nets.

### 5.1 Plans

Our plan execution model works in concert with a library of plans that the agent may be executing. The library is made up of *tasks*. Tasks may be *goals*, *methods* and *primitive actions*. **Caution:** Please do not read too much into the terms "goal" and "method." By these terms we mean nothing more than disjunctive and conjunctive nodes in the plan graph (respectively).

### 5.2 The Model

In this section we will walk through the explanation of observed actions. Working backward from the observation, we must explain how the observed action is chosen from the set of pending actions. Then we must explain how that set of actions is assembled. The explanation of the pending set covers the choice of subtasks as means to achieving higher level goals. It also covers the evolution of a plan execution episode over time.

Recall the discussion in the previous section — in or-der to take into account the fact that our agents may be carrying out multiple, interleaved, partially-ordered plans, we simply stipulate that the action happening at any time is one that is chosen from the set of pending actions. This action of choosing from the current pending set corresponds to the hollow arrows in Figure 5. We will define the pending set later.

(1)    happen($X$, $T+1$) ←
        pending($P$, $T$),
        $X \in P$,
        †pick($X$, $P$, $T+1$).

Rule 1 states that action $X$ happens at time $T+1$ if $X$ is in the set of pending actions and is picked from that set. pick($X$, $P$, $T$) is one of the distinguished hypothesis propositions with an associated probability. (In our presentation of the rules, we will mark hypotheses with a †.)

In our current system, we simply assume that all of the pending actions are equally likely. It would be simple to provide a more complex model of the choice of next action to execute, for example, some sort of exponentially-increasing probability as the action continues to be pending.

For example, two of the explanations for happen(open-p1, 1), from the plan database shown in Figure 1 are: (1) The pending set is {open-p1, check-temp} and open-p1 is picked; (2) The pending set is {open-p1, open-p2} and open-p1 is picked.

We define the pending set recursively, first providing a basis case, that allows us to explain the contents of the pending set at time zero. The process of choosing the initial pending set from the plan library is shown in Figure 5 through the narrow arrows. The pending set is selected from the set of primitive actions (leaves):

(2)    pending($L$, 0) ←
        leaves($Ls$),
        select_leaves($Ls, L$)).

The rule for select_leaves selects those leaves that are enabled.

The core of our model of plan execution is what it means for an action to be enabled. There are two reasons why an action may be enabled: either it is done for its own sake, or it is done as part of a parent plan.

(3)    enabled($A$, $T$) ← for_own_sake($A$, $T$).

(4)    enabled($A$, $T$) ←
        ¬ for_own_sake($A$, $T$),
        for_parent($A$, $T$).

The simple case is an action being done for its own case:

(5)    for_own_sake($A$, $T$) ←
        intendable($A$),
        †intended($A$, $T$).

The action $A$ must be one of the set of actions that may be done for their own sake. Whether or not a particular action meets this condition is indicated in the plan database. Note that there are no constraints on what *kind* of action may be done for its own sake; goals, methods and primitive actions are all eligible. In addition to it being possible that an action be done for its own sake, it must actually *be* done for its own sake. This is the second condition, that $A$ be intended, one of the hypotheses whose probability must be assessed (this will be discussed later).

The other reason why an action might be enabled is that it is being done as part of a larger plan:

(6)    for_parent($A$, $T$) ←
        parents($A$, $Ps$),
        $\exists P \in Ps \mid$ enabled_by($A$, $P$, $T$).

There are two ways a child action, $C$, can be enabled as part of a parent plan, $P$ (rules 7 and 10). First, the parent plan may be a method; a recipe that is a set of steps. In this case, in order for a particular action to be enabled, it must be a member of that set of steps, and all the preceding steps in that recipe must already have been done:

(7)    enabled_by($C$, $P$, $T$) ←
        and_node($P$),
        enabled($P$, $T$),
        preds($C, P$) = $\mathcal{P}$,
        $\forall p \in \mathcal{P}$ [ prev_done($p$, $T$) ].

The set of predecessors of a given node in a plan is defined in the plan library. For example, in Figure 1, the predecessor set of start-$O_2$-gen in gen-$O_2$ is { start-gen-B }.

The definition for prev_done for primitive actions is straightforward — a primitive action, $A$ is prev_done at $T$ iff happen($A, T'$) and $T' \leq T$. For composite actions, it is more complex. Rule 8 gives the condition for methods and 9 for goals.

(8)    prev_done($A$, $T+1$) ←
        and_node($A$),
        expansion($A$) = $\mathcal{A}$,
        $\forall a \in \mathcal{A}$ [ prev_done($a$, $T+1$) ].

(9)    prev_done($A$, $T+1$) ←
        or_node($A$),
        enabled($A$, $T+1$),
        †choose_expansion($M$, $A$),
        prev_done($M$, $T+1$).

The second way a child, $C$ can be enabled as part of a plan, $P$ is if $C$ is one of a set of methods for carrying out $P$. In this case, what is required is that the executing agent choose $C$ as the way she intends to do $P$:

(10)   enabled_by($C$, $P$, $T$) ←
         or_node($P$),
         choose_expansion($C$, $P$),
         enabled($P$, $T$).

Expressions of the form choose_expansion($C$, $P$) are hypotheses. In the examples we have worked, we have assumed that all of the methods for a given parent, $P$, are equally likely, however this may easily be changed. One need only supply choice probabilities. A more substantial limitation is that this approach assumes that only a *single* method will be chosen by the executing agent as his means of achieving $P$. This assumption could be relaxed, but only at the cost of a more cumbersome model of method choice. So far, we have not found this necessary.

So far, we have provided only the rules necessary for explaining the initial contents of the set of pending actions and hence, explaining the first action an agent takes. As a plan execution progresses, the contents of the pending set changes (this process is shown in the shaded arrows in Figure 5):

(11)   pending($P_{t+1}$, $T+1$) ←
         pending($P_t$, $T$),
         leaves($Ls$),
         progress($Ls$, $P_t$, $P_{t+1}$, $T+1$).

The progress rule assembles a new pending list, $P_{t+1}$ by considering the set of primitive actions, adding and subtracting appropriate actions. This is the most complex of the rules, involving several cases:

1. The most recent action is removed from the pending set;

2. All other previously pending actions remain pending;

3. Things that were not previously pending may be added.

The need for the rules to be mutually-exclusive and exhaustive makes the actual rules for progress somewhat cumbersome, so we have omitted them from the short version of this paper. The interesting case is the case of adding a primitive action to the pending set, Case 3. Primitive actions are added to the pending set if they have not already been done and if they are enabled:

(12)   add_me($A$, $T$) ←
         ¬ prev_done($A$, $T$),
         enabled($A$, $T$).

Recall the two ways an action can become enabled: if it is the chosen method for a newly-enabled parent (10) or when is part of a method that is being executed and the last of its predecessors has just been completed (7). Also, recall that the definitions of enabled are recursive, applying not only to primitive actions, but also to methods and goals.

### 5.3 Incorporating Context

So far, we have simply assumed that all plans have some fixed *a priori* probability of being chosen. We extend our model of plan execution to make plan adoption conditional on facts that obtain at the start of the episode. We add rules that make plan adoption conditional on some facts about the environment. We provide an example of this later in the paper.

### 5.4 Interacting with the Executing Agent

We wish to perform plan recognition in order to provide help to users. To do this properly, it is not enough to passively observe the process of plan execution. We must also be able to *intervene* into the process and predict the effects of our interventions. We would like to build intelligent systems that intervene by performing actions on the user's behalf.

Pearl (1994) provides a theoretical framework for causal interventions into processes described by dynamic Bayesian Networks like ours (Figure 5). The essence of his approach is that one can intervene by "clamping" a node in the network and cutting the causal links into that node.

Now we revise the model of plan execution so that we permit two explanations for an action occurring: either the observed agent carries out the action, or the system intervenes to perform an action. We apply Pearl's approach to our plan execution model by replacing rule (1) above with the following two rules:

(13)   happen($X, T+1$) ←intervene($X, T+1$).

and the following revision of the original rule (1), to make it mutually exclusive and exhaustive with (13):[1]

(14)   happen($X$, $T+1$) ←
         ¬ intervene($T+1$),
         pending($L$, $T$),
         member($X$, $L$),
         pick($X$, $L$, $T+1$).

Rule (13) indicates that an intervention makes the occurrance of action $X$ at $T+1$ independent of preceding events and the contents of the pending set. However, interventions *do* have causal effects on the future contents of the pending set and hence on what later actions are performed by the agent.

---

[1] PHA requires all rules be mutually-exclusive and exhaustive.

## 5.5 Probabilities

In order to make the model here a probabilistic model, we need only specify probabilities for a restricted set of hypotheses. These are as follows:

1. The probability that the agent will adopt a particular task *for its own sake*. These probabilities may either be simple, unconditional probabilities, or may be conditional on the agent's environment.

2. The probability that the agent will choose a particular method when attempting to achieve a goal.

3. The probability that the agent will choose a particular primitive action from a set of pending primitive actions.

These three parameters correspond to the three types of hypotheses, marked with a † in the rules above above.

The simple set of rules above, together with a plan library, provides a framework for abductive plan recognition. We may generate abductive proofs for the occurrence of observed action sequences, and from these explanations extract user intentions. In the following section, we provide several worked examples. These serve both to illustrate our model and to show how it differs from previous work in this area.

## 6 Plan recognition examples

In this section we provide examples that show how our model handles difficult plan recognition cases. In these examples, unless we state otherwise, we have assumed that all goals are equally likely and all methods are equally likely to be chosen by the agent. Most of these examples will be based on the plan library shown in Figure 1. We stress that these assumptions are *not* parts of our model — they are additional simplifications for the sake of clarity in presentation.

We abbreviate propositions like happen(open-p1, t) as open-p1$_t$, yielding expressions like $P($open-p1$_1)$: the probability that open-p1 happens at time 1, and $P($check-temp$_2|$open-p1$_1)$: the probability that check-temp happens at time 2 given that open-p1 happened at time 1. Likewise, we will use $P(intend($increase-power$)|$open-p1$_1)$ to express the system's posterior belief that the agent intends to increase-power given that open-p1 happened at time 1. Where it is possible without ambiguity, we will use $\omega$ to represent the observations, as in $P(intend($increase-power$)|\omega)$.

## 6.1 Interleaving plans for multiple goals

Consider the case where the agent executes action open-p1 and then action check-temp. Given this, the system can conclude that the agent must be pursuing raise-temp and one or both of increase-power and raise-$O_2$-level:

| Plan | $P(intend($Plan$|\omega))$ |
|---|---|
| raise-temp | 1.0 |
| increase-power | 0.6603 |
| raise-$O_2$-level | 0.6603 |

The system's beliefs about the multiple goals of the agent allow it to assign higher probability to actions that contribute to the agent's possible plans. For example, start-gen-B and raise-temp-set are the next steps in the respective plans. Therefore, the system believes one of these actions will almost certainly be the the next action executed:

$P($start-gen-B$_3|\omega)$ = .4748
$P($raise-temp-set$_3|\omega)$ = .4748

## 6.2 Partially ordered plans

The pending set provides our system the ability to handle partially ordered plans. Consider the case of actions open-p2, shutoff-X1 and shutoff-X2, which make up lower-power-use. In this domain, action open-p2 is the only predecessor for both shutoff-X1 and shutoff-X2. Thus once open-p2 has been executed both shutoff-X1 and shutoff-X2 are enabled and therefore are in the pending set until they are performed.

The rules for construction of the pending set guarantee that ordering constraints are enforced. For example, action start-gen-B cannot be added to the pending set until action open-p1 has been performed (see Figure 1).

## 6.3 Overloaded actions

In Figure 1, open-p1 is an example of an action that may be done as part of two plans, simultaneously. This possibility can be readily handled by our theory. Recall that in order to explain observations, we must find each pending set consistent with the evidence and consider its possible explanations. If we ask for the probability that open-p1 will be the first action performed, we must consider *all* of the possible goal sets the agent might be pursuing and *all* the possible expansions of those goals. As part of this process, we will find explanations where action open-p1 is used as part of two different methods: one for increase-power and another to raise-$O_2$-level.

### 6.4 Performing actions for their own sake

Let us compare two cases in which we observe **check-temp**. In the first case, we make the conventional assumption that **check-temp** will only be done as part of a plan to raising the temperature. In this case, we can prove that the agent must be executing **raise-temp**. If, on the other hand, we consider the possibility that the operator is idly checking the temperature, then we will conclude that the probabilities that the agent is pursuing **raise-temp** or just doing **check-temp**, is equal to their prior probabilities. If the probability of doing **check-temp** "for its own sake" is relatively high, then we will not necessarily conclude that the agent is trying to raising the temperature. If this plan recognition is being done as part of a system that might take actions to help the agent it is observing, one can see that getting this right is important.

### 6.5 Negative Evidence

Consider what happens if we were to see actions **open-p1** and **start-gen-B**. All else being equal, we can conclude that **increase-power** (by means of **gen-power**) and **raise-$O_2$-level** (via **gen-$O_2$**) are equally likely. But now consider what happens as we see action **check-temp**, and then action **raise-temp-set**. As time goes by and we see other actions, we become more and more confident that we have seen method **gen-power** and not simply the beginning of **gen-$O_2$**.

The use of the pending set in our framework provides a simple and elegant solution to this problem. Recall that an action happens if it is in the set of pending actions, and is picked from that set. In our framework, as we see the sequence of actions that does not contain **start-$O_2$-gen** grow, we will gradually become more and more certain that **start-$O_2$-gen** is not a member of the pending set. This will undermine our previous belief in hypothesis **raise-$O_2$-level** and reinforce our belief in **increase-power**. Table 1 shows how the conditional probabilities for this problem evolve over time.

### 6.6 Context

Our model allows us to take into account the state of the world when considering what goals the agent might be pursuing. This capability is essential in supporting users whose tasks are influenced by the state of the world, e.g. users that are controlling a manufacturing plant.

Figure 1 shows that the agent's intentions toward **increase-power** depend on the proposition **EVA-prep**. We assume for simplicity's sake that the re-

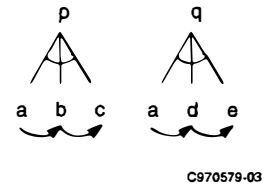

Figure 6: A plan library with two equally likely methods.

lationship between **EVA-prep** and **increase-power** is deterministic (this is not required by our formalism): if **EVA-prep** is true then the agent intends **increase-power** and if ¬**EVA-prep** then the agent will not intend **increase-power**. If $P(\textbf{EVA-prep}) = 0.5$ then all three plans are equally likely a priori: $P(\textbf{increase-power}) = P(\textbf{raise-}O_2\textbf{-level}) = P(\textbf{raise-temp}) = 0.5$

The probabilities of the initial actions for **increase-power** are $P(\textbf{open-p1}_1) = .2864$ and $P(\textbf{open-p2}_1) = .1458$. However, if the system knows that **EVA-prep** is true, then its belief that the agent is pursuing **increase-power** shifts up to one and the probability of any action that contributes to this plan increases. Thus $P(\textbf{open-p1}_1|\textbf{EVA-prep}) = .3854$ and $P(\textbf{open-p2}_1|\textbf{EVA-prep}) = .2916$.

### 6.7 Interventions

Consider the set of actions given in Figure 6. Suppose the agent is observed performing action **a**. The only plans that are consistent with performing **a** are **p** and **q** and these are equally likely. Therefore, the system concludes that the agent is doing either **p** or **q** with equal probability: $P(intend(\textbf{p})|\textbf{a}_1) = P(intend(\textbf{q})|\textbf{a}_1) = .6666$. It further concludes that the actions **b** and **d** are equally likely to appear next: $P(\textbf{b}_2|\textbf{a}_1) = P(\textbf{d}_2|\textbf{a}_1) = .5$.

Suppose the agent were to perform **b**. Since **p** is the only reason to do **b**, the system can safely infer that the agent intends **p**. This also tends to explain away **a**, lowering the probability of **q**: $P(intend(\textbf{q})|\textbf{a}_1,\textbf{b}_2) = .3333$. The execution of **b** also enables **c** and provides evidence that it will be next: $P(\textbf{c}_3|\textbf{a}_1,\textbf{b}_2) = .8333$.

Now consider what would happen if, rather than the agent performing **b**, the system had done it for her (we write this as $I(\textbf{b}_2)$). The plan recognition algorithm's only treats **b** as evidence for the agent's pursuit of **p** if the agent herself does it. If the system performs **b**, then the system's belief in the probability of the agent pursuing **p** does not change: $P(intend(\textbf{p})|\textbf{a}_1) = P(intend(\textbf{p})|\textbf{a}_1, I(\textbf{b}_2)) = .6666$

However, the execution of **b** by the system does licence

| Time | Proposition | Probability |
|---|---|---|
| 2 | $P(intend(\text{increase-power})\|\text{open-p1}_1,\text{start-gen-B}_2)$ | .6414 |
|   | $P(intend(\text{raise-}O_2\text{-level})\|\text{open-p1}_1,\text{start-gen-B}_2)$ | .6414 |
| 3 | $P(intend(\text{increase-power})\|\text{open-p1}_1,\text{start-gen-B}_2,\text{check-temp}_3)$ | .7416 |
|   | $P(intend(\text{raise-}O_2\text{-level})\|\text{open-p1}_1,\text{start-gen-B}_2,\text{check-temp}_3)$ | .4832 |
| 4 | $P(intend(\text{increase-power})\|\text{open-p1}_1,\text{start-gen-B}_2,\text{check-temp}_3,\text{raise-temp-set}_4)$ | .8282 |
|   | $P(intend(\text{raise-}O_2\text{-level})\|\text{open-p1}_1,\text{start-gen-B}_2,\text{check-temp}_3,\text{raise-temp-set}_4)$ | .3128 |

Table 1: Results showing the cumulative impact of negative evidence.

*some* inferences. For example, since b has been executed c is now enabled, and therefore will be added to the pending set. Since the system's execution of b does not change the system's beliefs about the goals the agent is persuing, the system believes that c is as likely to be the next action as b was before its execution: $P(\mathbf{b}_2|\mathbf{a}_1) = P(\mathbf{c}_3|\mathbf{a}_1, I(\mathbf{b}_2)) = .5$.

## 7 Implementation

Our plan recognition theory has been implemented computationally. The implementation centers around a PHA abductive theorem prover. Our theorem-prover is based on David Poole's (Poole 1993a) but, it differs substantially in search strategy. All examples mentioned in this paper have been tested and work in our implementation.

The rules used by the program differ from those presented earlier only in the treatment of quantification and negation, neither of which are interpreted by the theorem prover. We expect to make our theorem prover and rule set publicly available.

## 8 Concluding Remarks

In this paper we have presented a general model of plan recognition based on probabilistic abductive logic. The centerpiece of the model is a simple formal theory of plan execution. In comparison with previous work on plan recognition, our theory better handles sequences of actions generated by interleaved, partially-ordered plans. Our theory can accommodate information about the context in which plans are adopted. Finally, unlike previous theories, ours can incorporate interventions into the process of plan execution.

In future work, we would like to address some of the limitations of our model. One limitation is that in this model the agent does not interact with its surroundings. The actions done by the agent and the interventions do not change the world state, because the world state is not represented. We are now working on a more elaborate plan execution model that takes this into account.

**Acknowledgements** Thanks to David Poole for his PHA interpreter and help working with it.

**References**

Charniak, E., and Goldman, R. P. 1993. A Bayesian model of plan recognition. *Artificial Intelligence* 64(1):53–79.

Charniak, E., and McDermott, D. 1985. *Introduction to Artificial Intelligence*. Reading, MA: Addison Wesley.

Cohen, P. R.; Perrault, C. R.; and Allen, J. F. 1981. Beyond question answering. In Lehnert, W., and Ringle, M., eds., *Strategies for Natural Language Processing*. Hillsdale, NJ: Lawrence Erlbaum Associates. 245–274.

de Mantaras, R. L., and Poole, D., eds. 1994. *Uncertainty in Artificial Intelligence, Proceedings of the Tenth Conference*. Morgan Kaufmann.

Kautz, H., and Allen, J. F. 1986. Generalized plan recognition. In *Proceedings of the Fifth National Conference on Artificial Intelligence*, 32–38.

Pearl, J. 1994. A probabilistic calculus of actions. In de Mantaras and Poole (1994), 454–462.

Poole, D. 1993a. Logic programming, abduction and probability: a top-down anytime algorithm for stimating prior and posterior probabilities. *New Generation Computing* 11(3–4):377–400.

Poole, D. 1993b. Probabilistic horn abduction and Bayesian networks. *Artificial Intelligence* 64:81–129.

Pynadath, D. V., and Wellman, M. P. 1997. Generalized queries on probabilistic context-free grammars. To appear in IEEE *PAMI*.

Vilain, M. 1990. Getting serious about parsing plans: A grammatical analysis of plan recognition. In *Proceedings of the Eighth National Conference on Artificial Intelligence*, 190–197. Cambridge, MA: MIT Press.

Wellman, M. P., and Pynadath, D. V. 1997. Plan recognition under uncertainty. Unpublished web page.